# Capturing and Recognizing Objects Appearance Employing Eigenspace

M. Ashrafuzzaman[1], M. Masudur Rahman[2] M M A Hashem[3]

[1]Shakti Engineering Ltd., Lalmatia, Dhaka, Bangladesh, asubitk@hotmail.com

[2]Department of Control Engineering, Kyushu Institute of Technology, Japan

[3]Dept. of Computer Science and Engineering, Bangladesh Institute of Technology (BIT), Khulna, Bangladesh

Emails: asubitk@hotmail.com, csebitk@bttb.net.bd, rahman@is.cntl.kyutech.ac.jp

**Abstract**

This paper presents a method of capturing objects appearances from its environment and it also describes how to recognize unknown appearances creating an eigenspace. This representation and recognition can be done automatically taking objects various appearances by using robotic vision from a defined environment. This technique also allows extracting objects from some sort of complicated scenes. In this case, some of object appearances are taken with defined occlusions and eigenspaces are created by accepting both of non-occluded and occluded appearances together. Eigenspace is constructed successfully every times when a new object appears, and various appearances accumulated gradually. A sequence of appearances is generated from its accumulated shapes, which is used for recognition of the unknown objects appearances. Various objects environments are shown in the experiment to capture objects appearances and experimental results show effectiveness of the proposed approach.

## 1. Introduction

Automatic accumulation and recognition of objects appearances using an intelligent vision system has been paid much attention to the modern researchers. In the past years, we have seen an extensive use of appearance-based matching that is based on capturing the appearances of objects under various conditions. A set of learning images is generally encoded via principal component analysis [5] and represented as an eigenspace. It should be noted that an objects rotation gives many appearances according to viewing orientations about its axis and the present study, as a primary step, is limited to deal with a few objects considering some sort of occlusions. In general, appearances are to be captured using camera and a particular learning image set is encoded by a single eigenspace with choosing its reasonable dimension. In a similar way, a set of appearances makes a consequence group of eigenspaces in a defined space. Various objects can be accumulated using different eigenspaces. Unknown images can be tested after projecting it onto this eigenspace, and by recognizing the nearest neighbor.

Object accumulation and recognition is one of the basic actions for robots during household helps. It is very difficult to learn and recognize automatically of an object's character/shape in order to hold/carry it. Besides, it is more difficult to handle objects with critical shapes, complex figures or grouped objects [3] using available method to the day. In some other cases, robots have to recognize variable scenes where location, position or pose of objects can be varied.

In this study, we allow some sort of occlusions, which is the key factor of this method. A few of occluded appearances are included into non-occluded appearances for producing an eigenspace. The obtained eigenspace, which includes some of occluded appearances, is able to recognize unknown similar appearances successfully.

Appearance of objects is one of the key factors to identify particular object and determine their exact poses overcoming the preceding problems. Based on this vision, the robot can observe an object like a human vision and we need not any partial specifications. The present paper is somewhat different with its application from the principle viewpoints [1]. Recently, development of IDC (Intelligent Data Carrier), RFID (Radio Frequency Identification) [2] and powerful computer has impressed to think about this idea that is used for distributed information management [4]. By using such devices, the robot can share the acquired knowledge through the devices. The system learns about the object appearance, create its eigenspace and proceed to hold it after recognizing. The process construct the eigenspace from successively acquired object appearances. If the object is known, the robot goes to recognize it just observing one of the viewpoints; otherwise it learns the unknown object taking appearances for further recognition.

The proposed method is explained briefly in section 2. Experimental activities and their result are described in section 3. In section 4, the study is concluded with discussing some important issues.

## 2. Creating eigenspace using image appearances

Let us consider $n$ objects in an environment and $m$ appearances (including occluded appearances) can be taken from each object. If $X$ is considered for an image set,

$$X = \{x_{1,1}, x_{1,2}, ..., x_{1,m}, x_{2,m}..., x_{nm}\}, \quad (1)$$

where, $x$ indicates an individual appearance having $i,j$ size. A covariance $Q$ matrix is calculated as follows;

$$Q = XX^T \quad (2)$$

After that, we can obtain eigenvalues $\lambda_i$ with its corresponding eigenvectors $e_i$ of the matrix $Q$ using an eigen equation. The dimensional space is compressed by the chosen $k$ ($k$=1,2…) eigenvectors using PCA algorithm and eigenvectors $e_i$ ($i$=1,2,…,$k$) corresponding to the largest $k$ eigenvalues are obtained. If we consider an eigenspace $E$ for a particular object, the eigenspace can be denoted as;

$$E(e_1, e_2, ..., e_k)^T x_{ij}. \quad (3)$$

The constructed eigenspace is useful to store many appearances in a small memory space, since it projects images to a lower dimensional eigenspace keeping principal features. The study has not described detail of the procedures, these can be found in the literature [1].

## 3. Experiment & Result

In this section, an extensive experimental activity is discussed. In this particular study, we have taken four different objects in different image situations. It is noted that the background effect have not considered here. However, some sort of occlusions is considered in this study.

A particular object is targeted and taken its snap with possible different viewpoints using a camera. **Figure 1** shows the objects taken into considerations. Four different objects, i.e., a key holder, a mobile set, a pencil box and a stapler machine are employed in this experiment. In this particular study, 10 different appearances/images are taken from its 100-degree (10-degree interval) environment from an object. We define an environment as an object can be placed in different situations such as dispersed situation or in a group. **Figure 2** shows some of objects placed in various environments. In this figure (left), appearances of objects can be extracted and recognized easily and objects recognition is possible by using this technique. However, the extraction is really difficult (right) and some sort of occluded objects are taken from this environment for making a combined eigenspace.

The taken appearances are sampled into image portion and these images are further proceeded for creating its own eigenspace. When an eigenspace is created for a particular object, the procedures do continue for surrounding objects. When a new object appears in the defined environment, it is compared and recognized according to the conventional method. If it is not available, the new object's appearances are also captured according to the earlier described way. Successive eigenspace is constructed from successively acquired objects. A constructed eigenspace is shown in **Figure 3** where 10 different appearances of the mobile set are stored. The points indicate the different image locations as it has been taken within 100-degree rotation. In this figure, only an occluded appearance is included for investigating the effect of occlusions problem. The individual points indicate the transformed extracted appearances in eigenspace. The axes show three largest eigenvectors labeled by Eigenvector-e1, Eigenvector-e2 and Eigenvector-e3, respectively. Once we obtain the eigenspace of a particular object, the system is continued until all images are captured and stored into eigenspace.

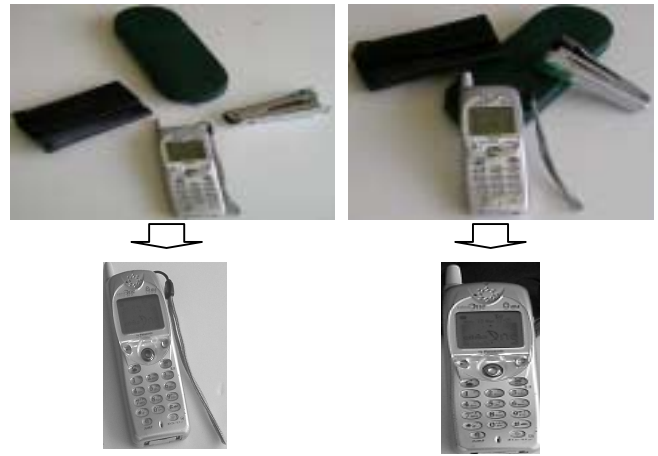

**Figure 2**. Objects placed in various environments. Appearances of objects can be extracted and recognized (left) easily and objects extraction is really difficult (right).

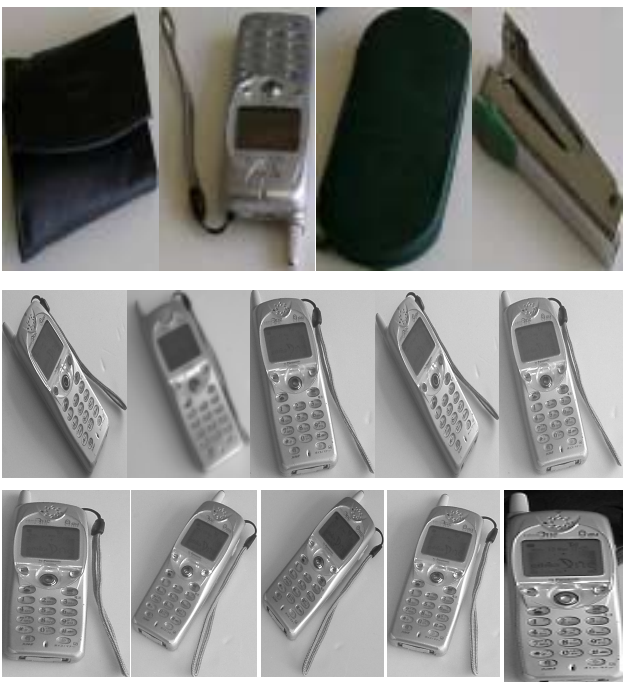

**Figure 1**. Some of objects taken in this experiment (upper row) and appearances of an object that are used for producing eigenspace.

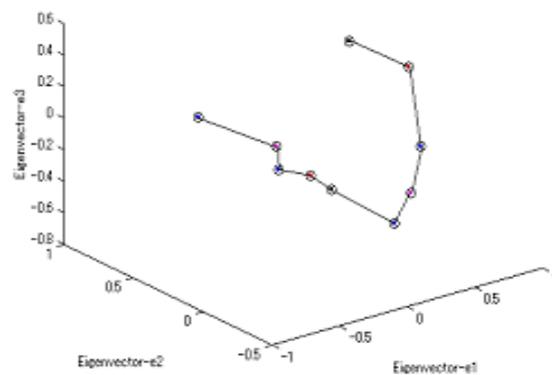

**Figure 3**. An eigenspace. Ten different appearances are represented in an eigenspace. Only three prominent dimensions are displayed here.

Therefore, we have all of objects eigenspace and if we want to recognize an unknown appearance, the image is projected onto the eigenspace. The location of the nearest appearance indicates the exact position of the projected image representing an unknown appearance. The recognition rate $r$ is calculated using the following formula;

$$r = m/P, \qquad (4)$$

where $P$ is the number of projected images for every subject and $m$ the number of successfully recognized images.

In the experimental result, we have successfully extracted the objects appearances and stored the images creating their eigenspace. In the same time, we have recognized the unknown objects appearances using the respective eigenspace. As described, we have employed an object for creating a basic eigenspace with considering an occluded image. Besides, other three objects are employed for testing the recognition rates. In this test sample, each object has taken one occluded appearance. In this study, we have received average of 96% recognition rates.

## 4. Conclusion

The proposed method has shown a better advancement in terms of recognition rates for recognizing objects in occlusion. Consideration of occlusion problems and the images are extracted from a defined environment instead of taking just images of objects make the difference to the original method. In addition, the present approach directly interacts with the different environments as if a robot can work independently. We have successfully acquired the appearances objects by constructing their eigenspaces. The acquired appearances are stored into the computer memory instead of intelligent agent and various appearances of the object are accumulated using eigenspace constructions. Due to some important advantages, this technique can be of interest to the all researchers.

In general, automatic object accumulation and recognition in the environment may not be possible if the object is occluded by any means, objects in under cluttered background, change their figures after destruction. We have shown that our method can overcome some sort of occlusions problems. However, more considerations of occlusions problems, background effects and changes of illuminations were out of scope of this paper. Besides, we have used only four models for proving the effectiveness of the proposed approach, which cannot be a sufficient for a recommending a model. The consequences of study can be solved these issues.

## 5. References

1. H. Murase, S. K. Nayar, "Visual learning and recognition of 3-D objects from appearance", *Int. J. Computer Vision*, 14, 5, 39-50 (1995).
2. Thomas Von Numers et. al., "An Intelligent data carrier system for local communication between cooperative multiple mobile robots and environment", *2nd IFAC Conf. on Intelligent Autonomous Vehicle*, pp.-366-371(1995).
3. M. Masudur Rahman, Seiji Ishikawa, "Appearance-Based Human Postures Recognition Overcoming Dress Effect", *Int. Conf. on Computational Intelligent for Modeling, Control and Automation*, pp.271-278 (2001).
4. Yasushi Mae, et. al., "Acquisition and accumulation of objects appearances using successive eigenspace construction", *The 5th Asian Conf. on Computer Vision*, pp.1-6 (2002).
5. Gonzalez, R. C., Wintz, P., *Digital Image Processing*, Addison-Wesley Publishing Company Limited (1986).